\documentclass[journal]{IEEEtran}
\usepackage{amsmath,amsfonts}
\usepackage{algorithmic}
\usepackage{array}
\usepackage[caption=false,font=normalsize,labelfont=sf,textfont=sf]{subfig}
\usepackage{textcomp}
\usepackage{stfloats}
\usepackage{url}
\usepackage{verbatim}
\usepackage{graphicx}
\def\BibTeX{{\rm B\kern-.05em{\sc i\kern-.025em b}\kern-.08em
    T\kern-.1667em\lower.7ex\hbox{E}\kern-.125emX}}
\usepackage{balance}
\begin{document}
\title{Towards Human-Like Interactive Speech Recognition with Agentic Correction and Semantic Evaluation}


\author{
Zixuan Jiang,
Yanqiao Zhu,
Peng Wang,
Qinyuan Chen, 
Xinjian Zhao, 
Xipeng Qiu, 
Wupeng Wang, 
Zhifu Gao, 
Xiangang Li, 
Kai Yu, 
and Xie Chen%
\thanks{Zixuan Jiang, Yanqiao Zhu, and Peng Wang contributed equally to this work. (Corresponding author: Xie Chen.)}%
\thanks{Zixuan Jiang is with the College of Artificial Intelligence, Xi'an Jiaotong University, Xi'an 710049, China (e-mail: andrewjiang@stu.xjtu.edu.cn). This work was conducted during his internship at X-LANCE Lab, School of Electronic Information and Electrical Engineering, Shanghai Jiao Tong University, Shanghai 200240, China.}%
\thanks{Yanqiao Zhu, Kai Yu, and Xie Chen are with X-LANCE Lab, School of Electronic Information and Electrical Engineering, Shanghai Jiao Tong University, Shanghai 200240, China (e-mail: 1850432206@sjtu.edu.cn; kai.yu@sjtu.edu.cn; chenxie95@sjtu.edu.cn). This work was conducted during Yanqiao Zhu's internship at Tongyi Fun Team, Alibaba Group}%
\thanks{Peng Wang is with The Chinese University of Hong Kong, Shenzhen, Shenzhen 518172, China (e-mail: pengwang0104@gmail.com).}%
\thanks{Qinyuan Chen, Xingjian Zhao, Xipeng Qiu are with Fudan University, Shanghai 200433, China (e-mail: chengqy21@m.fudan.edu.cn; zhaoxj24@m.fudan.edu.cn; xpqiu@fudan.edu.cn).}%
\thanks{Wupeng Wang, Zhifu Gao, Xiangang Li are with Tongyi Fun Team, Alibaba Group, Hangzhou 310030, China (e-mail: wangwupeng.wwp@alibaba-inc.com; Zhifu.gzf@alibaba-inc.com; lixiangang.lxg@alibaba-inc.com).}%
}

\markboth{JOURNAL OF LATEX CLASS FILES, VOL. 14, NO. 8, AUGUST 2021}%
{Towards Human-Like Interactive Speech Recognition With Agentic Correction and Semantic Evaluation}

\maketitle

\begin{abstract}
Automatic speech recognition (ASR) is a core component of human--computer interaction and an increasingly important front-end for LLM-based assistants and agents. However, most current ASR systems still follow a single-pass paradigm, which is poorly aligned with human communication, where misunderstandings are resolved through iterative clarification and refinement. This mismatch makes it difficult to correct meaning-critical errors once they occur. Meanwhile, token-level metrics such as WER or CER cannot adequately reflect such a problem. To address these limitations, we formulate \emph{Interactive ASR} as a multi-turn refinement task and propose \textbf{Agentic ASR}, a closed-loop framework that combines a single-pass ASR front-end with semantic correction, intent routing, and reasoning-based editing. We further introduce the \textbf{Sentence-level Semantic Error Rate} ($S^2ER$), an LLM-based semantic evaluation metric, together with an \textbf{Interactive Simulation System} for scalable and reproducible benchmarking. Experiments on multilingual, named-entity-intensive, and code-switching benchmarks show that iterative interaction consistently reduces semantic errors, with much larger gains in $S^2ER$ than in conventional token-level metrics. Human--AI alignment and ablation studies further validate the reliability of the semantic judge and the robustness of the proposed framework. The code is available at: \url{https://interactiveasr.github.io/} and the live demo is available at \url{https://i-asr.sjtuxlance.com/}
\end{abstract}


\section{Introduction}

\begin{figure*}[t]
    \centering
    \includegraphics[width=\textwidth]{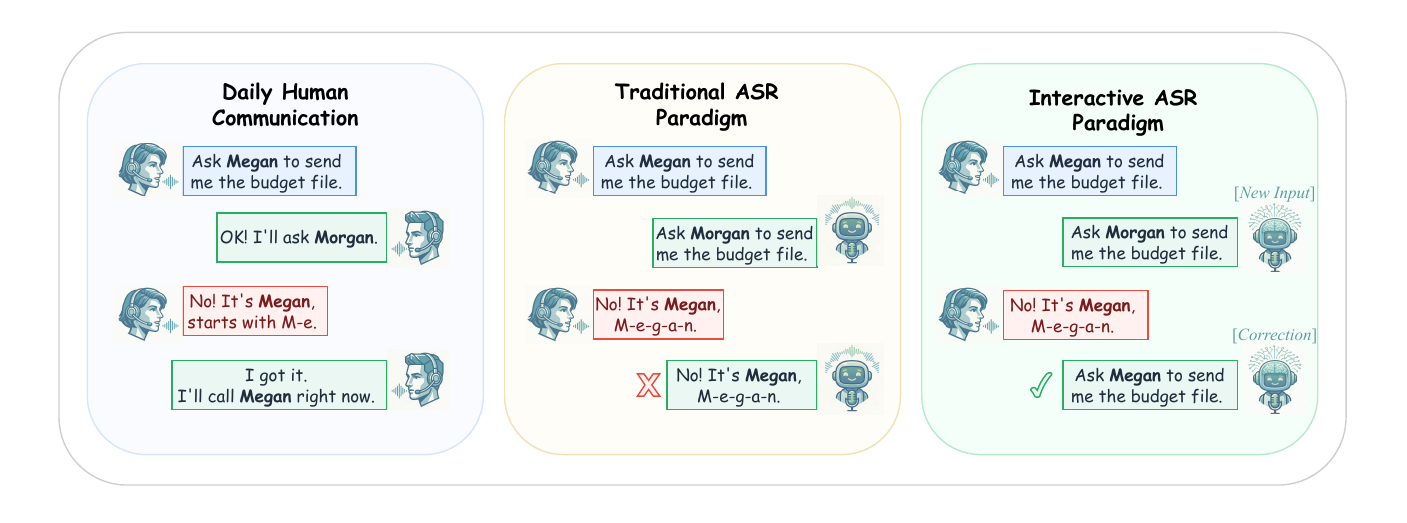}
    \caption{
\textbf{Comparison between daily human communication, the traditional ASR paradigm, and the proposed Agentic ASR paradigm. }
In natural conversations, misunderstandings can be progressively corrected through multi-turn interactions. 
In contrast, conventional ASR systems operate in a one-shot, open-loop manner, where recognition errors (e.g., confusing ``Megan'' with ``Morgan'') cannot be effectively corrected once produced. 
The Agentic ASR paradigm introduces a closed-loop mechanism that incorporates user feedback, enabling iterative refinement of transcription results and more accurate understanding.}
    \label{fig:teaser}
\end{figure*}

\IEEEPARstart{A}{utomatic} speech recognition (ASR)\cite{asr} has become a core component of human--computer interaction, especially as speech increasingly serves as the input interface for large language model (LLM)-based assistants and agents \cite{wang2023slm, qwen3asr_shi}. Recent progress in end-to-end ASR, including attention-based encoder--decoder architectures \cite{aed_chan} and large-scale weakly supervised speech models \cite{aed_whisper_radford}, together with scaling in model capacity and training data, has substantially improved recognition accuracy. However, modern ASR systems are still largely designed as single-pass transcription engines. This one-shot design departs from natural human communication, where misunderstandings are routinely resolved through feedback and repair.

As illustrated in Fig.~\ref{fig:teaser}, the mismatch becomes most evident when errors affect meaning-critical content. Cognitive and conversation studies suggest that human dialogue is grounded through iterative confirmation and self-repair \cite{clark1991grounding, schegloff1977selfcorrection}. For example, if someone mishears \emph{Megan} as \emph{Morgan}, the speaker will usually offer a brief correction, and the listener will adjust their understanding accordingly. Conventional ASR systems, by contrast, usually interpret a follow-up utterance as new input rather than as a correction to an earlier hypothesis. Once an error is produced, the system therefore has little ability to revise it within the same interaction. This limitation is particularly problematic in scenarios involving named entities, spelling clarification, accented speech, background noise, or code-switching, where ambiguity is common and interactive repair is often necessary \cite{Jannet2015howtoevaluateasroutputforner}.

The mismatch is not only procedural but also evaluative. Standard ASR metrics such as Word Error Rate (WER) \cite{wang2003word} and Character Error Rate (CER) \cite{Jelinek1997StatisticalMF} quantify token-level mismatch, but they do not explicitly distinguish minor surface deviations from errors that change the intended meaning. Prior studies on semantic ASR evaluation, including Semantic WER \cite{roy2021semantic} and SemDist \cite{kim2021semantic}, have shown that semantic preservation is not always well captured by conventional token-level measures. This gap becomes more consequential in interactive and agent-oriented settings, where the practical impact of an error depends less on its local form than on whether it distorts named entities, user intent, or other task-critical content.

These observations motivate a rethinking of ASR task from two complementary perspectives: \emph{mechanism} and \emph{evaluation}. From the mechanism side, ASR should move beyond one-shot prediction and support iterative correction under user feedback. From the evaluation side, ASR should be assessed not only by token accuracy but also by sentence-level semantic equivalence. To this end, we define the \textbf{Interactive ASR} task, in which a system progressively refines its hypothesis through multi-turn interaction. We then propose \textbf{Agentic ASR}, and introduce the \textbf{Sentence-level Semantic Error Rate ($S^2ER$)} together with an \textbf{Interactive Simulation System (ISS)} for scalable and reproducible multi-turn evaluation.

Our main contributions are summarized as follows:
\begin{itemize}
    \item \textbf{Interactive ASR task formulation.} We define \textbf{Interactive ASR} as a stateful multi-turn transcription task, formalizing ASR as iterative refinement under user feedback instead of independent single-pass decoding.
    \item \textbf{Agentic solution for interactive correction.} We propose \textbf{Agentic ASR}, a closed-loop framework with semantic correction, intent routing, and reasoning-based correction to progressively repair meaning-critical recognition errors.
    \item \textbf{Semantic evaluation system for Interactive ASR.} We establish a dedicated evaluation scheme with \textbf{$S^2ER$} and \textbf{ISS}, and verify its reliability and effectiveness through human--AI alignment and multi-benchmark experiments.
\end{itemize}

\section{Related Work}

\subsection{ASR Metrics}

Word Error Rate (WER) remains the standard metric for ASR, but it assigns uniform costs to all tokens and edit operations, and therefore does not reflect the unequal semantic impact of different recognition errors.

To address this limitation, prior work has extended edit-distance metrics with semantic sensitivity. \textbf{WWER} \cite{shichiri-etal-2008-automatic} introduces context-dependent weights over words and operations. \textbf{NE-WER} \cite{Jannet2015howtoevaluateasroutputforner} and \textbf{Semantic WER} \cite{roy2021semantic} place greater emphasis on content-bearing tokens, especially named entities. \textbf{H\_eval} \cite{sasindran2023hevalnewhybridevaluation} further combines lexical edits with semantic-distance signals to balance token accuracy and meaning preservation.

Another line of work evaluates semantic similarity more directly. \textbf{BERTScore} \cite{zhang2020bertscoreevaluatingtextgeneration} measures contextual token-level similarity, while \textbf{SemDist} \cite{kim2021semantic} uses sentence-level embedding distance. More recently, LLM-based evaluation has also been explored: \textbf{LASER} \cite{parulekar2025laser} grades error severity with an LLM, and \textbf{Answer Error Rate} \cite{pulikodan2025approachmeasuringperformanceautomatic} evaluates ASR through downstream QA behavior.

Different from these approaches, our \textbf{$S^2ER$} uses a binary functional criterion: whether the hypothesis preserves sufficient meaning for correct intent execution in an interactive setting. This design explicitly targets interaction success rather than fine-grained lexical similarity or constructed downstream proxy scores.

\subsection{Human-Feedback-Based ASR Approaches}

Human feedback has long been used to improve ASR outputs, mostly through explicit correction interfaces. Early systems support multimodal selection from N-best candidates \cite{suhm2001multimodal} or touch-based text correction in voice-typing interfaces \cite{CHI-2012-KumarPL}. Another practical strategy is acoustic respeaking, where users re-utter misrecognized content \cite{sperber2013efficient}.

Recent work also uses user corrections to improve ASR models themselves. \textbf{The Gift of Feedback} \cite{zhou2023giftfeedbackimprovingasr} leverages on-device corrections via federated learning, mainly for model adaptation to long-tail terms. However, this line primarily treats feedback as training supervision, rather than as online guidance for resolving errors during the ongoing interaction.

In parallel, NLP agent frameworks such as \textbf{ReAct} \cite{yao2022react} show that natural-language feedback can guide iterative reasoning and revision. Building on this paradigm, our framework allows users to provide open-form natural-language feedback to repair ASR errors \emph{in place}, bridging rigid correction interfaces and language-driven interactive refinement.

\subsection{LLM as a Judge}
\label{ssec:llm-as-a-judge}

LLM-as-a-judge methods have been widely studied in machine translation \cite{kocmi2023largelanguagemodelsstateoftheart}, dialogue \cite{zheng2023judging}, and summarization \cite{liu2023gevalnlgevaluationusing}, due to strong semantic understanding and comparative reasoning.

In ASR, \cite{Liu2024EvaluatingSR} evaluates transcription quality using LLM-based semantic representations and shows better alignment with downstream task success than purely lexical metrics, although hidden-state-based scoring is less interpretable. LATTEScore \cite{10447177} makes this direction more explicit by casting ASR semantic assessment as a binary semantic-preservation decision. Our \textbf{$S^2ER$} follows the same binary-evaluation principle, and further focuses on interactive-ASR applicability by aligning the judgment target with intent-preserving usability.

A separate but important issue is judgment stability. LLMBar \cite{zeng2024evaluatinglargelanguagemodels} shows that rubrics, examples, and swap-and-synthesis reduce variance, and \cite{shi2025judgingjudgessystematicstudy} reports positional sensitivity in long-context evaluation while showing that multi-sample aggregation improves robustness. Our evaluation protocol adopts these stability practices (e.g., order swapping and multi-round aggregation) to improve reliability of LLM-based ASR semantic judgment.

\section{Agentic ASR}
\label{sec:proposed_method}

\subsection{Task Formulation}
\label{ssec:task_formulation}

A conventional ASR system transcribes each speech input in a single pass:
\begin{equation}
Y = \mathrm{ASR}(I),
\end{equation}
where the output is determined only by the current acoustic signal. Once this output is produced, later user feedback is not explicitly incorporated into the same decoding process.

In contrast, Interactive ASR formulates transcription as a stateful multi-turn refinement process. Let $Y_{[:t-1]}=\{Y_0,\ldots,Y_{t-1}\}$ denote the transcription history up to turn $t-1$. At the initial turn, the system receives the first speech input $I_0$ and produces
\begin{equation}
Y_0 = \mathrm{InteractiveASR}(\emptyset, I_0),
\end{equation}
where $\emptyset$ denotes the absence of prior context. For each subsequent turn $t>0$, the system updates the transcription state by conditioning on both the new speech input and the interaction history:
\begin{equation}
Y_t = \mathrm{InteractiveASR}(Y_{[:t-1]}, I_t).
\end{equation}

This formulation casts Interactive ASR as a recurrent state-update problem rather than independent one-shot recognition. The key difference from conventional ASR is that each turn explicitly uses historical context, enabling progressive repair of meaning-critical errors through interaction.

\subsection{Agentic ASR Framework}
\label{ssec:interactive_asr_framework}

\begin{figure}[!t]
    \centering
    \includegraphics[width=\columnwidth]{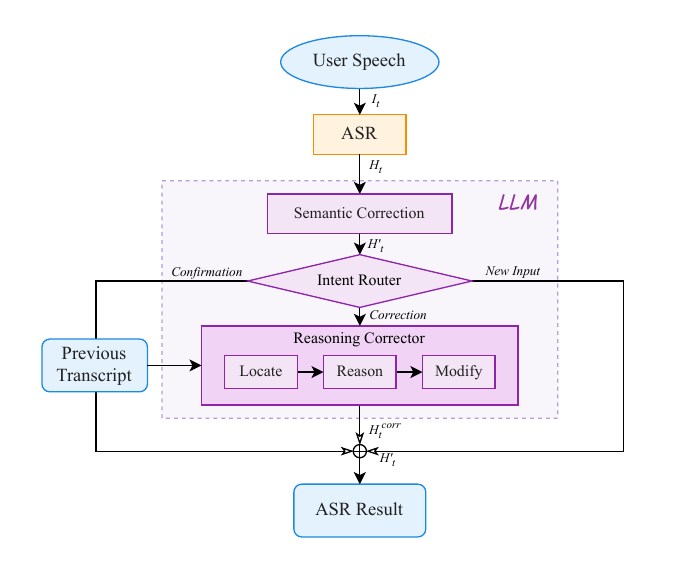}
    \caption{Agentic ASR framework. At turn $t$, an ASR front-end first produces a hypothesis $H_t$ from user speech input $I_t$. An LLM module then performs semantic correction and intent routing into three intent types: \textit{confirmation}, \textit{new input}, and \textit{correction}. For \textit{correction} intents, a structured \textit{Locate--Reason--Modify} pipeline identifies the editable span, infers the intended edit from instruction and history, and applies the edit to update the transcription state.}
    \label{fig:iasr-framework}
\end{figure}

To instantiate the formulation above, we propose \textbf{Agentic ASR}, an architecture that combines a single-pass ASR front-end with an LLM-based reasoning-and-editing module. At turn $t$, the framework takes user speech $I_t$ and transcription history $Y_{[:t-1]}$ as input, and outputs an updated transcription state $Y_t$. As shown in Fig.~\ref{fig:iasr-framework}, the update is carried out through three stages: semantic correction, intent routing, and reasoning-based correction.

Given the current speech input $I_t$, the ASR front-end first generates a textual hypothesis
\begin{equation}
H_t = \mathrm{ASR}(I_t).
\end{equation}
At a high level, the LLM module refines this hypothesis conditioned on transcription history:
\begin{equation}
Y_t = \mathrm{LLM}(H_t, Y_{[:t-1]}; \mathcal{P}_{\mathrm{refine}}),
\end{equation}
where $\mathcal{P}_{\mathrm{refine}}$ specifies the refinement protocol. To improve interpretability and controllability, we explicitly decompose this refinement process into the following stages.

\paragraph{Semantic correction.}
The first stage maps the raw ASR hypothesis to an explicit, semantically coherent instruction:
\begin{equation}
H_t' = \mathrm{SemanticCorrection}(H_t, Y_{[:t-1]}).
\end{equation}
This stage is necessary because user correction utterances are themselves recognized by the ASR front-end and may therefore contain semantic inconsistencies. By leveraging interaction history $Y_{[:t-1]}$, the semantic-correction module rewrites $H_t$ into an explicit and executable instruction $H_t'$ that is consistent with the current context.

\paragraph{Intent routing.}
The corrected instruction $H_t'$ is then classified into one of three intent categories:
\begin{equation}
c_t \in \{\mathrm{confirmation},\ \mathrm{new\ input},\ \mathrm{correction}\}.
\end{equation}
Here, \emph{confirmation} indicates that the user accepts the current state, \emph{new input} indicates that the current utterance should be interpreted as newly transcribed content, and \emph{correction} indicates that the utterance should edit the existing transcription state. The state update rule is then defined as follows:
\begin{equation}
Y_t =
\begin{cases}
Y_{t-1}, & c_t = \mathrm{confirmation},\\
H_t', & c_t = \mathrm{new\ input},\\
H_t^{\mathrm{corr}}, & c_t = \mathrm{correction}.
\end{cases}
\label{eq:agent_update}
\end{equation}

\paragraph{Reasoning-based correction.}
When $c_t=\mathrm{correction}$, the framework invokes a reasoning module to revise the transcription state according to the current instruction:
\begin{equation}
H_t^{\mathrm{corr}} = \mathrm{ReasoningCorrector}(H_t', Y_{[:t-1]}; \mathcal{P}_{\mathrm{corr}}),
\end{equation}
where $\mathcal{P}_{\mathrm{corr}}$ is the correction-oriented reasoning prompt. Rather than relying on an unconstrained one-step rewrite, we decompose correction into three explicit operations:
\begin{equation}
\mathrm{ReasoningCorrector}
=
\mathrm{Modify}
\circ
\mathrm{Reason}
\circ
\mathrm{Locate}.
\end{equation}
Specifically, \emph{Locate} identifies the span to edit in the current history, \emph{Reason} infers the intended modification from $H_t'$ and the interaction context, and \emph{Modify} applies that edit to produce the updated state $H_t^{\mathrm{corr}}$. This decomposition makes the correction process more controllable and better aligned with how users naturally provide partial repair instructions.

\section{Sentence-level Semantic Error Rate ($S^2ER$)}

\subsection{Definition}
\label{ssec:sser}

To evaluate meaning preservation at the utterance level, we introduce the \emph{Sentence-level Semantic Error Rate} ($S^2ER$):
\begin{equation}
S^2ER=\frac{1}{N}\sum_{i=1}^{N}\left(1-\hat{z}_i\right),
\end{equation}
where $\hat{z}_i\in\{0,1\}$ indicates whether ASR hypothesis $Y_i$ is semantically equivalent to reference $Y_{GT,i}$. Specifically, $\hat{z}_i=1$ denotes semantic equivalence, and $\hat{z}_i=0$ denotes a meaning-critical error. Therefore, $S^2ER$ measures the proportion of utterances whose transcriptions fail to preserve intended meaning.

The judgment criterion is task-oriented. The judge focuses on whether main intent and key meaning-bearing content are preserved, especially proper nouns, named entities, and other task-critical information, while ignoring non-semantic variations such as disfluencies, filler words, and punctuation. Because the target is binary semantic equivalence, we use a concise prompt specification rather than an exhaustive rule list.

To improve robustness and reduce order sensitivity, we adopt a three-round bidirectional voting protocol, following the LLM-as-a-Judge stability practices discussed in Section~\ref{ssec:llm-as-a-judge}. In round $r$, the judge is queried twice with reversed input order:
\begin{equation}
\begin{aligned}
z_{i,r}^{(1)} &= \mathrm{LLM}_{\mathrm{judge}}(Y_i, Y_{GT,i}; \mathcal{P}_{\mathrm{judge}}), \\
z_{i,r}^{(2)} &= \mathrm{LLM}_{\mathrm{judge}}(Y_{GT,i}, Y_i; \mathcal{P}_{\mathrm{judge}}),
\end{aligned}
\end{equation}
where $z_{i,r}^{(1)}, z_{i,r}^{(2)} \in \{0,1\}$. A round is counted as positive only when both decisions indicate equivalence. The final label is obtained by majority voting across the three rounds:
\begin{equation}
\hat{z}_i=\mathbf{1}\!\left(\sum_{r=1}^{3}\bigl(z_{i,r}^{(1)} \land z_{i,r}^{(2)}\bigr)\ge 2\right).
\end{equation}
This protocol mitigates input-order bias and improves label stability.

\subsection{$S^2ER$ versus token-level metrics}

Token-level metrics such as WER and CER measure surface-form mismatch, but they do not distinguish semantically negligible errors from meaning-critical ones. This limitation is especially problematic in interactive and agent-oriented ASR, where downstream success depends primarily on whether the transcription preserves the user intent and key entities, rather than on exact lexical matching.

\begin{figure}[!t]
    \centering
    \includegraphics[width=0.95\linewidth]{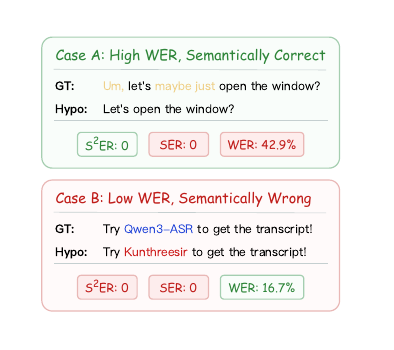}
    \caption{Two illustrative cases comparing $S^2ER$ with token-level metrics. In Case A, several mismatches involve only filler or discourse words, leading to high WER but preserved meaning. In Case B, a single local substitution corrupts a key entity, yielding lower WER but a semantic failure.}
    \label{fig:s2er_vs_token}
\end{figure}

Figure~\ref{fig:s2er_vs_token} shows two representative cases. In Case A, the reference utterance is ``Um, let's maybe just open the window?'' while the hypothesis is ``Let's open the window?'' This yields a relatively high WER of $42.9\%$, because several mismatched tokens are discourse or filler words. However, the main intent remains unchanged, and the utterance is still fully usable for downstream execution; it should therefore be judged as semantically correct, corresponding to $\hat{z}=1$ and contributing no error to $S^2ER$. In contrast, Case B changes only one local span: ``Try Qwen3-ASR to get the transcript!'' becomes ``Try Kunthreesir to get the transcript!'' Here the WER is only $16.7\%$, yet the substituted token is a meaning-critical entity. In an agentic setting, such an error may directly break model selection, retrieval, or tool routing, even though the lexical deviation is small.

These examples show that $S^2ER$ captures functional usability more directly than token-level metrics. A sentence with many non-essential token errors may still preserve the intended meaning and incur no semantic error, whereas a sentence with only one local substitution may become unusable if that substitution corrupts a key entity or intent-bearing word. This difference also explains why, in later experiments, interaction often yields much larger gains in $S^2ER$ than in WER, CER, or MER: iterative correction mainly repairs meaning-critical errors rather than merely polishing local surface mismatches.

\subsection{Interactive Simulation System}

\begin{figure}[!t]
    \centering
    \includegraphics[width=\linewidth]{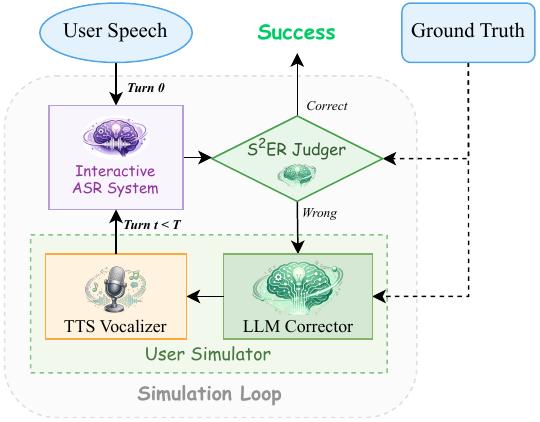}
    \caption{Interactive Simulation System (ISS) for automatic multi-round evaluation of Interactive ASR. At each round, the $S^2ER$ Judger evaluates semantic equivalence between the current transcription and ground truth; if equivalence is not achieved, the User Simulator produces corrective spoken feedback for the next round.}
    \label{fig:ISF}
\end{figure}

Large-scale human-in-the-loop evaluation of Interactive ASR is expensive and hard to reproduce. To enable scalable and repeatable benchmarking, we design an \emph{Interactive Simulation System} (ISS), illustrated in Fig.~\ref{fig:ISF}. ISS simulates multi-round user--system interaction and uses the $S^2ER$ Judger in Section~\ref{ssec:sser} as the round-wise semantic stopping criterion.

For sample $i$, ISS starts from initial user speech $X_{i,0}$ and ground-truth transcription $Y_{i,\mathrm{GT}}$. At each round, the evaluated Interactive ASR system outputs a transcription, which is evaluated by the $S^2ER$ Judger for semantic equivalence. If equivalence is not reached, a User Simulator generates the next corrective spoken instruction. The interaction stops once equivalence is achieved or a predefined maximum number of rounds is reached.

\subsubsection{$S^2ER$ Judger in ISS}

At round $t$, given current transcription $Y_{i,t}$, the $S^2ER$ Judger outputs a binary semantic-equivalence label
\begin{equation}
    \hat{z}_{i,t}\in\{0,1\},
\end{equation}
by applying the same protocol in Section~\ref{ssec:sser} to pair $(Y_{i,t}, Y_{i,\mathrm{GT}})$. Here, $\hat{z}_{i,t}=1$ means semantic equivalence.

When $\hat{z}_{i,t}=1$, interaction for sample $i$ terminates, and remaining rounds are marked successful:
\begin{equation}
    \hat{z}_{i,[t:T]} := 1,
\end{equation}
where $T$ is the predefined maximum number of interaction rounds. Otherwise, $Y_{i,t}$ is forwarded to the User Simulator to generate next-round user input.

\subsubsection{User Simulator}

The User Simulator consists of an LLM-based corrector and a TTS vocalizer:
\begin{equation}
    X_{i,t+1} = \mathrm{TTS}\big(\mathrm{LLM}(Y_{i,t}, Y_{i,\mathrm{GT}}, \mathcal{P}_{\mathrm{corr}})\big),
\end{equation}
where $X_{i,t+1}$ is simulated user speech at round $t+1$. The LLM corrector compares $Y_{i,t}$ with $Y_{i,\mathrm{GT}}$, identifies the key semantic discrepancy, and generates a concise correction instruction. The TTS vocalizer converts this instruction into speech and feeds it back to the evaluated Interactive ASR system.

For a dataset with $N$ samples, round-wise $S^2ER$ at round $t$ is
\begin{equation}
    S^2ER_t = \frac{1}{N}\sum_{i=1}^{N}\left(1-\hat{z}_{i,t}\right).
\end{equation}
This quantity measures the proportion of samples that remain semantically incorrect after round $t$. The same simulation framework can be extended to other metrics (e.g., WER and CER) by replacing the $S^2ER$ Judger with the corresponding metric-specific evaluator.

\section{Experiments}
\label{sec:experiment}

In this section, we evaluate both the proposed Agentic ASR framework and the semantic evaluation protocol built around $S^2ER$. Section~\ref{ssec:experiment_setup} introduces the model configuration and evaluation datasets. Section~\ref{ssec:human_ai_alignment} verifies that the LLM judge aligns with human semantic judgments. Section~\ref{ssec:Mainresult} analyzes the main performance trends across multilingual, named entity and code-switching \cite{chen2022aishellnernamedentityrecognition}\cite{agro2025codeswitchingendtoendautomaticspeech} benchmarks, and Section~\ref{ssec:Ablation_Study} examines the effects of ASR backbone choice, LLM scale, and judge strategy. 

\subsection{Experiment Setup}
\label{ssec:experiment_setup}

\subsubsection{Model Configuration}
\label{sssec:config}

Unless otherwise specified, \textbf{Qwen3-ASR-1.7B}~\cite{qwen3asr_shi} is used as the backbone ASR model to generate initial transcription hypotheses. \textbf{Qwen3-32B}~\cite{yang2025qwen3} is used as the unified LLM for three components: the Reasoning LLM in Agentic ASR, the Correction Generator in the User Simulator, and the Semantic Judge in $S^2ER$ evaluation. For speech synthesis in the simulation framework, we employ \textbf{Index-TTS-1.5}~\cite{deng2025indextts}. The reference audio of each sample is used as the acoustic prompt to preserve speaker consistency across interaction turns.

\subsubsection{Evaluation Datasets}
\label{sssec:datasets}

To evaluate robustness and generalization under multilingual, named-entity-intensive, and code-switching conditions, experiments are conducted on evaluation splits from representative benchmarks in three categories.

\paragraph{Multilingual speech.}
We use \textbf{GigaSpeech Test}~\cite{chen2021gigaspeech} and \textbf{WenetSpeech Test\_Net}~\cite{zhang2022wenetspeech} to evaluate English and Mandarin open-domain ASR performance, respectively.

\paragraph{Named-entity-intensive speech.}
We construct \textbf{AISHELL-NER Dev$^\dagger$} and \textbf{AISHELL-NER Test$^\dagger$} by filtering the AISHELL-1 development and test splits with AISHELL-NER annotations~\cite{bu2017aishell1opensourcemandarinspeech,chen2022aishellnernamedentityrecognition}, retaining only utterances containing named entities.

\paragraph{Code-switching speech.}
We use \textbf{ASRU2019 Test}~\cite{shi2020asru} and \textbf{CS-Dialogue Test$^\dagger$}~\cite{zhou2025csdialogue104hourdatasetspontaneous} for Mandarin--English code-switching evaluation. The latter is constructed by selecting code-switching utterances from the original CS-Dialogue corpus.

\subsection{Human--AI Alignment Study}
\label{ssec:human_ai_alignment}

To validate the reliability of $S^2ER$, we conduct a human--AI correlation study following prior work on \textit{Spoken Language Assessment} \cite{zechner2019automated, biswas2021automated} and LLM-based evaluation. We sample 40 utterances each from GigaSpeech (English), WenetSpeech (Chinese), and ASRU2019 (code-switching), yielding a validation set of 120 examples that covers the main linguistic conditions considered in this paper.

Semantic consistency between each ASR output and its reference transcript is independently annotated by 25 non-expert annotators and 5 domain experts using a binary protocol, where 1 denotes semantic consistency and 0 otherwise. For each sample, the averaged human score serves as the reference target.

The same validation set is then evaluated by the LLM Judger described in Section~\ref{ssec:sser}. We compute Pearson correlation coefficients ($r$) \cite{pearson1895} between the LLM judgments, expert judgments, and the human reference scores, and repeat the LLM evaluation over five runs to assess stability.

\begin{table}[htb]
\caption{Correlation between the LLM Judger, experts, and human reference scores across datasets.}
\label{tab:human_ai_corr}
\centering
\begin{tabular}{lcccc}
\hline
\textbf{Dataset} & \textbf{LLM $r$} & \textbf{Std} & \textbf{Expert $r$} & \textbf{Diff} \\
\hline
GigaSpeech   & 0.8914 & 0.0291 & 0.8534 & \textbf{+0.0380} \\
WenetSpeech  & 0.8280 & 0.0252 & 0.8086 & \textbf{+0.0194} \\
ASRU2019     & 0.9031 & 0.0250 & 0.8871 & \textbf{+0.0160} \\
\hline
\end{tabular}

\vspace{2pt}
\footnotesize{LLM $r$: mean Pearson correlation over five runs; 
Std: standard deviation; 
Expert $r$: mean Pearson correlation of expert evaluations; 
Diff: difference between LLM $r$ and Expert $r$.}
\end{table}

Table~\ref{tab:human_ai_corr} shows that the LLM Judger tracks human semantic judgments reliably across all three datasets. Its correlation with the human reference scores remains above 0.8 throughout, and it is consistently slightly higher than that of the domain experts. The standard deviations across five runs are also small, indicating that the judgment protocol is stable rather than sensitive to sampling noise or prompt randomness. Taken together, these results support the use of $S^2ER$ as a reliable semantic evaluation signal for ASR.

\subsection{Main Results}
\label{ssec:Mainresult}

\begin{figure}[!t]
    \centering
    \includegraphics[width=\linewidth]{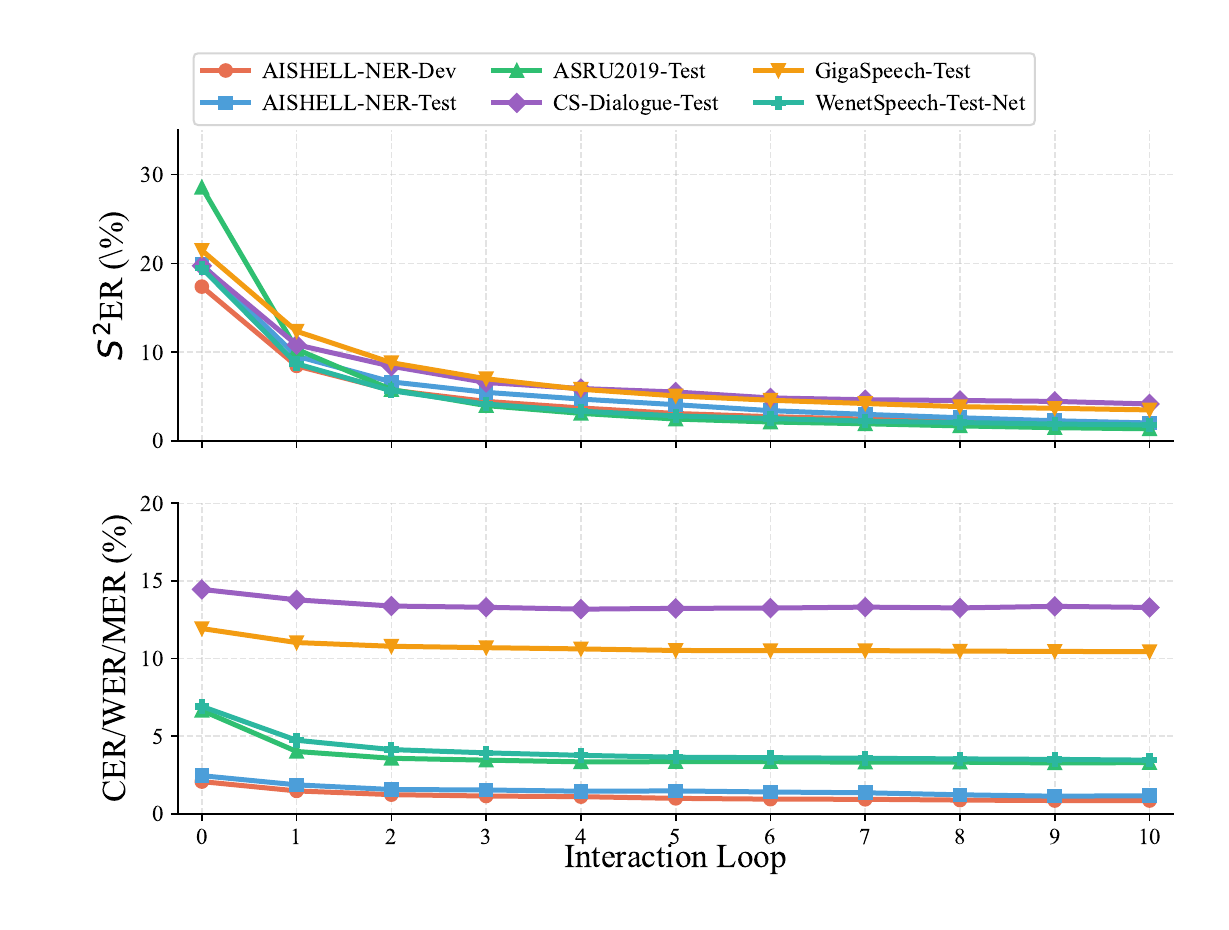}
    \caption{Performance trends of the proposed Agentic ASR framework from Loop 0 to Loop 10. The upper panel shows $S^2ER$ across all benchmarks, while the lower panel reports conventional error rates (WER/CER/MER). $S^2ER$ decreases consistently with more interaction, with the largest gains typically observed in the first few loops, whereas conventional token-level metrics improve more gradually.}
    \label{fig:main_result_curve}
\end{figure}

Table~\ref{tab:main_results} reports representative checkpoints of the proposed Agentic ASR framework, while Fig.~\ref{fig:main_result_curve} shows the full trajectories from Loop 0 to Loop 10. The central finding is that multi-turn interaction consistently improves transcription quality across all benchmarks, with the largest gains appearing at the semantic level.

\begin{table*}[t]
\centering
\caption{Main results of the proposed Agentic ASR framework under different numbers of interaction loops. Representative checkpoints (Loops 0, 1, 3, and 10) are reported, while the full trajectories from Loop 0 to 10 are shown in Fig.~\ref{fig:main_result_curve}. Increasing the interaction budget consistently improves semantic correctness across multilingual, named-entity-intensive, and code-switching benchmarks.}
\setlength{\tabcolsep}{4pt}
\renewcommand{\arraystretch}{1.15}
\begin{tabular}{c|cc|cc|cc|cc|cc|cc}
\hline
\textbf{Loop} 
& \multicolumn{2}{c|}{\textbf{GigaSpeech Test}} 
& \multicolumn{2}{c|}{\textbf{WenetSpeech Test\_Net}} 
& \multicolumn{2}{c|}{\textbf{AISHELL-NER Dev$^\dagger$}} 
& \multicolumn{2}{c|}{\textbf{AISHELL-NER Test$^\dagger$}} 
& \multicolumn{2}{c|}{\textbf{CS-Dialogue Test$^\dagger$}} 
& \multicolumn{2}{c}{\textbf{ASRU2019 Test}} \\
\cline{2-13}
& \textbf{$S^2ER$} & \textbf{WER}
& \textbf{$S^2ER$} & \textbf{CER}
& \textbf{$S^2ER$} & \textbf{NER}
& \textbf{$S^2ER$} & \textbf{NER}
& \textbf{$S^2ER$} & \textbf{MER}
& \textbf{$S^2ER$} & \textbf{MER} \\
\hline
0  & 21.47\% & 11.92\% & 19.46\% & 6.91\% & 17.38\% & 2.07\% & 19.91\% & 2.45\% & 19.73\% & 14.44\% & 28.57\% & 6.65\% \\
1  & 12.35\% & 11.02\% & 8.69\%  & 4.73\% & 8.45\%  & 1.47\% & 9.55\%  & 1.86\% & 10.83\% & 13.77\% & 10.32\% & 4.01\% \\
3  & 7.00\%  & 10.69\% & 4.15\%  & 3.92\% & 4.45\%  & 1.14\% & 5.47\%  & 1.53\% & 6.58\%  & 13.29\% & 3.98\%  & 3.45\% \\
10 & \textbf{3.49\%} & \textbf{10.43\%} & \textbf{1.80\%} & \textbf{3.45\%} & \textbf{1.97\%} & \textbf{0.85\%} & \textbf{2.02\%} & \textbf{1.16\%} & \textbf{4.16\%} & \textbf{13.28\%} & \textbf{1.36\%} & \textbf{3.29\%} \\
\hline
\end{tabular}
\label{tab:main_results}
\vspace{2pt}
\begin{flushleft}
\footnotesize
\textit{Note:} $S^2ER$ denotes the proposed Sentence-level Semantic Error Rate. WER, CER, NER, and MER denote Word Error Rate, Character Error Rate, Named-Entity Error Rate, and Mixed Error Rate, respectively. WER is reported for English speech, CER for Mandarin speech, NER for named-entity-intensive subsets, and MER for code-switching benchmarks.
\end{flushleft}
\end{table*}

$S^2ER$ decreases monotonically with additional interaction on every benchmark. Table~\ref{tab:main_results} shows large one-step gains immediately after the first feedback turn, and Fig.~\ref{fig:main_result_curve} shows that the improvement continues through later loops without reversal. For example, $S^2ER$ on GigaSpeech Test drops from 21.47\% at Loop 0 to 12.35\% after only one interaction and reaches 3.49\% by Loop 10, while ASRU2019 Test improves from 28.57\% to 10.32\% after one loop and to 1.36\% by Loop 10. This pattern indicates that even limited feedback resolves a substantial portion of meaning-critical errors, while additional rounds further refine the remaining difficult cases.

Most of the semantic benefit arrives early. Figure~\ref{fig:main_result_curve} shows a steep drop in $S^2ER$ during the first few interaction rounds, followed by gradually diminishing but still positive returns. This shape is practically important: the framework recovers most semantic errors with a small interaction budget, rather than requiring many rounds before becoming useful.

The benefit is strongest precisely in the scenarios where semantic repair matters most. On the named-entity-intensive subsets, the final $S^2ER$ falls to around 2\%, suggesting that the framework is effective at repairing errors involving proper nouns and other high-value content. On code-switching benchmarks, the improvements remain substantial, although CS-Dialogue Test$^\dagger$ improves less than ASRU2019 Test. This gap likely reflects the greater difficulty of spontaneous conversational code-switching, but the consistent downward trend still shows that iterative semantic refinement remains effective under realistic interaction conditions.

The semantic gains are markedly larger than the improvements observed in conventional token-level metrics. Table~\ref{tab:main_results} and Fig.~\ref{fig:main_result_curve} show that $S^2ER$ drops much more sharply than WER, CER, or MER across datasets, indicating that the proposed framework mainly repairs meaning-critical errors rather than merely polishing local token mismatches. This discrepancy is precisely why a semantic metric is necessary: surface-form metrics alone would substantially understate the practical value of interaction.

Overall, the main results support two conclusions. First, Agentic ASR improves transcription quality through iterative interaction across multilingual, named-entity-intensive, and code-switching settings. Second, the gains are fundamentally semantic in nature, which confirms the need to evaluate interactive ASR with a meaning-oriented metric such as $S^2ER$ rather than with token-level measures alone.

\subsection{Ablation Study}
\label{ssec:Ablation_Study}

\subsubsection{Different Base ASR Model}
\label{sssec:asr-model-ablation}

\begin{figure*}[!t]
    \centering
    \includegraphics[width=\textwidth]{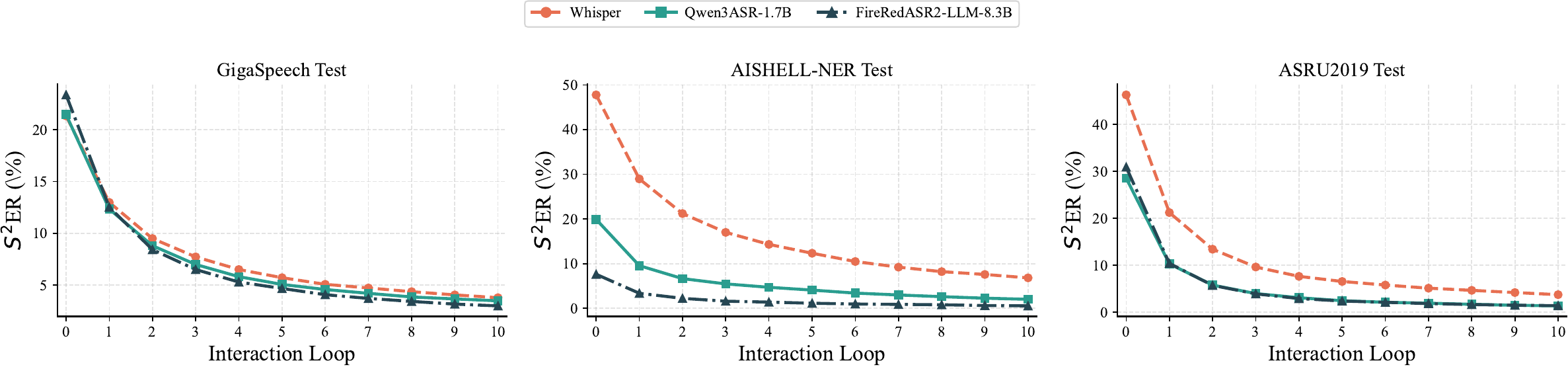}
    \caption{Ablation on different base ASR models under the same proposed Agentic ASR framework. Three representative shared benchmarks are shown: GigaSpeech Test for single-language recognition, AISHELL-NER Test$^\dagger$ for named-entity-intensive recognition, and ASRU2019 Test for code-switching recognition. All backbones benefit from iterative interaction, and even the weak Whisper model reaches much lower final $S^2ER$ after multiple loops.}
    \label{fig:asr_model_ablation}
\end{figure*}

We first examine whether Agentic ASR depends on a particular ASR backbone by replacing the default \textbf{Qwen3ASR-1.7B}\cite{qwen3asr_shi} with two alternatives: \textbf{FireRedASR2-LLM-8.3B}\cite{fireredasr_xu}, a larger and stronger recognizer, and \textbf{Whisper}\cite{aed_whisper_radford}, a substantially weaker small model. All other components of the interactive pipeline are kept unchanged. Figure~\ref{fig:asr_model_ablation} illustrates three representative benchmarks covering single-language, named-entity-intensive, and code-switching recognition.

Agentic ASR remains effective across a wide range of ASR backbones, including very weak ones. Figure~\ref{fig:asr_model_ablation} shows the same qualitative pattern on all three benchmarks: regardless of whether the recognizer is strong, moderate, or weak, $S^2ER$ decreases steadily as interaction proceeds. This consistency indicates that the gain comes from the interaction mechanism itself rather than from a narrow compatibility with one particular ASR architecture.

This point is most clearly demonstrated by the weakest backbone, Whisper. Although its starting point is much worse than that of the other models, multi-turn interaction still produces large and practically meaningful improvements. On AISHELL-NER Test$^\dagger$ and ASRU2019 Test, Whisper begins with Loop-0 $S^2ER$ values of 47.77\% and 46.32\%, respectively, yet iterative interaction reduces them to 6.82\% and 3.75\% by the final loop; on GigaSpeech Test, it reaches 3.79\%. In other words, even when the initial transcription is poor enough that nearly half of the utterances contain meaning-critical semantic errors, the proposed framework can still recover most of these errors after several rounds of user feedback. This result is important because it shows that Agentic ASR is not merely polishing already strong hypotheses; it can substantially lift weak ASR systems toward usable semantic accuracy through interaction alone.

At the same time, the base recognizer still affects the final error floor. Stronger backbones generally retain an advantage after the full interaction budget is exhausted: for example, the final $S^2ER$ on AISHELL-NER Test$^\dagger$ is 0.55\% for FireRedASR2-LLM-8.3B, 2.02\% for Qwen3ASR-1.7B, and 6.82\% for Whisper. The correct interpretation is therefore twofold. First, better initial ASR quality still improves the ultimate ceiling. Second, and more importantly for this ablation, weaker ASR models are not disqualified from the interactive setting; with sufficient multi-turn correction, they can still achieve strong final semantic performance. Agentic ASR is thus both robust to backbone choice and complementary to future improvements in base ASR quality.

\subsubsection{Size of LLM Reasoner}
\label{sssec:size-refiner-ablation}

We study the effect of LLM scale by replacing both the \textbf{Reasoning LLM} in Agentic ASR and the \textbf{Correction Generator} in the User Simulator with \textbf{Qwen3-8B}, while keeping all other components unchanged.

\begin{figure}[t]
    \centering
    \includegraphics[width=\linewidth]{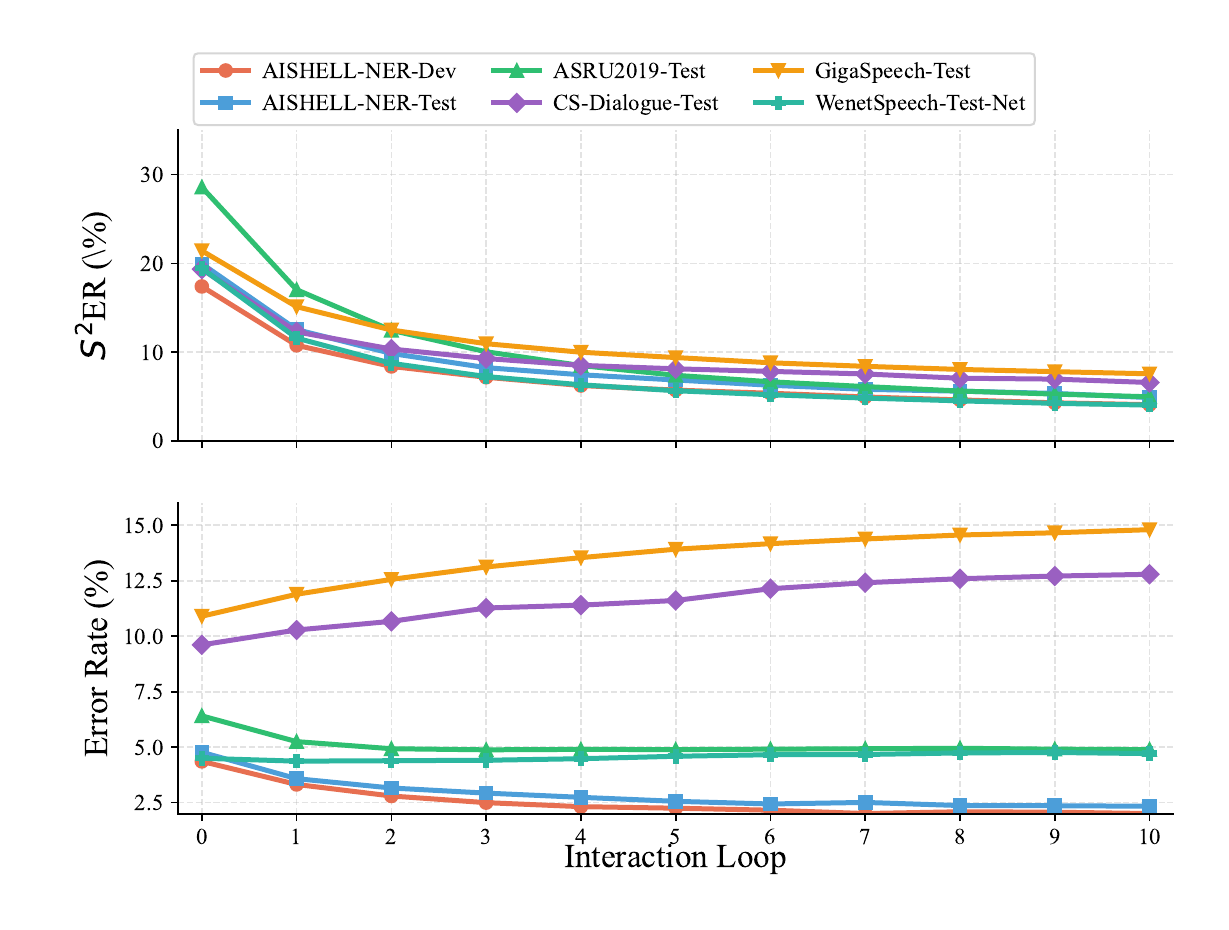}
    \caption{Ablation on the size of the LLM reasoner. Both the Reasoning LLM in Agentic ASR and the Correction Generator in the User Simulator are replaced with Qwen3-8B. The 8B variant still benefits from iterative interaction, but remains consistently worse than the default 32B setting.}
    \label{fig:size_refiner_ablation}
\end{figure}

A smaller LLM is still sufficient to preserve the basic benefit of interaction. Figure~\ref{fig:size_refiner_ablation} shows that the 8B variant maintains a monotonic decrease in $S^2ER$ on all benchmarks, indicating that even a compact model can support meaningful multi-turn correction and user-feedback simulation.

LLM scale substantially affects the quality of the final correction. Table~\ref{tab:size_refiner_ablation} shows that the 8B model is worse than the default 32B model on every dataset at Loop 10, with absolute gaps ranging from 2.11 to 4.07 points. The gap is therefore not an isolated failure case; it is a systematic penalty that appears across multilingual, named-entity-intensive, and code-switching settings.

\begin{table}[!t]
\centering
\caption{Effect of reducing the LLM reasoner from Qwen3-32B to Qwen3-8B. Final $S^2ER$ at Loop 10 is reported.}
\setlength{\tabcolsep}{5pt}
\renewcommand{\arraystretch}{1.1}
\begin{tabular}{lccc}
\hline
\textbf{Dataset} & \textbf{Qwen3-32B} & \textbf{Qwen3-8B} & \textbf{$\Delta$} \\
\hline
GigaSpeech Test            & 3.49\% & 7.56\% & +4.07\% \\
WenetSpeech Test\_Net      & 1.80\% & 4.03\% & +2.23\% \\
AISHELL-NER Dev$^\dagger$  & 1.97\% & 4.08\% & +2.11\% \\
AISHELL-NER Test$^\dagger$ & 2.02\% & 4.88\% & +2.86\% \\
CS-Dialogue Test$^\dagger$ & 4.16\% & 6.58\% & +2.42\% \\
ASRU2019 Test              & 1.36\% & 4.97\% & +3.61\% \\
\hline
\end{tabular}
\label{tab:size_refiner_ablation}
\end{table}

A more revealing difference is that conventional error rates become less stable under Qwen3-8B and may even worsen on some benchmarks. For example, WER on GigaSpeech Test and MER on CS-Dialogue Test$^\dagger$ increase as interaction proceeds. This divergence suggests that the smaller model often preserves enough global meaning to improve $S^2ER$, but is less reliable when the task requires precise and tightly constrained local edits.

We attribute this behavior to weaker instruction following and poorer edit precision in the 8B model. Compared with the 32B reasoner, it is more likely to generate ambiguous correction instructions, misidentify the target span, or rewrite beyond the intended scope. As a result, the interaction loop can still repair sentence-level semantics, but may simultaneously introduce local token mismatches, which explains the degradation of WER/CER/MER on some datasets.

Overall, this ablation shows that Qwen3-8B remains useful and preserves the main advantage of iterative interaction, but stronger LLMs provide more reliable correction and a lower final semantic error floor. LLM capability therefore does not determine whether Agentic ASR works, but it strongly influences how cleanly and how far the correction process can go.

\subsubsection{LLM-as-a-Judge Strategy}
\label{sssec:llm-as-a-judge-ablation}

We further examine the effect of repeated voting in the LLM-as-a-Judge protocol used for $S^2ER$. Specifically, we compare four variants: \textit{single}, which performs one bidirectional judgment, and \textit{majority-$K$} with $K\in\{3,5,7\}$, where the bidirectional judgment is repeated for $K$ rounds and the final label is determined by majority voting. The purpose of this ablation is to test whether repeated voting improves agreement with human judgments, and whether additional rounds remain worthwhile once the voting cost is taken into account.

\begin{figure}[t]
    \centering
    \includegraphics[width=\linewidth]{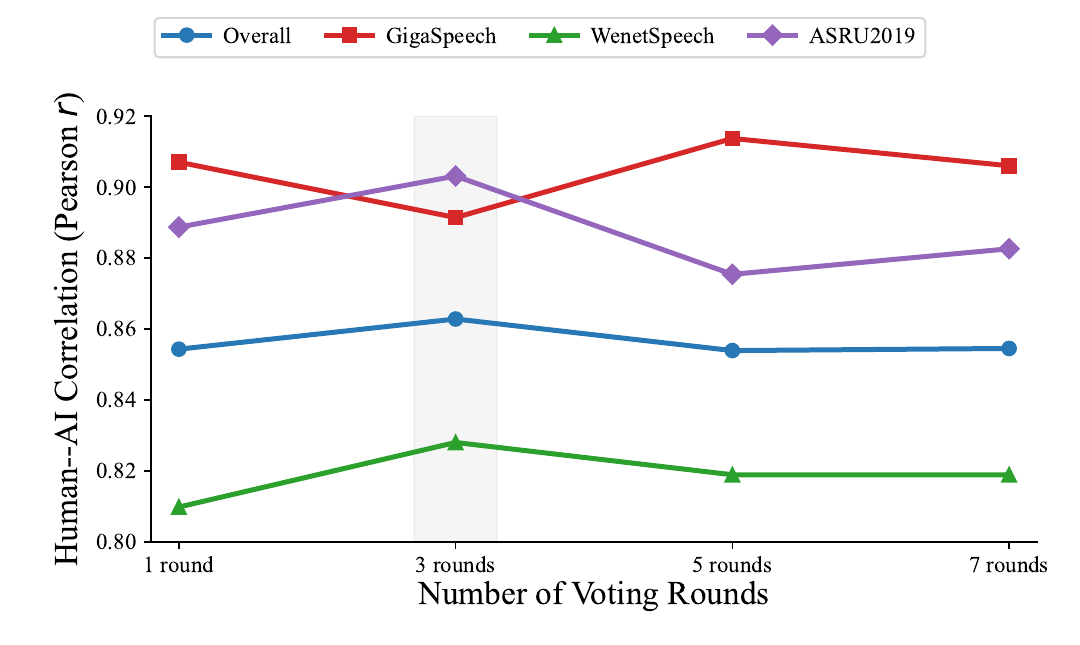}
    \caption{Mean Pearson correlation with human reference scores under different LLM-as-a-Judge voting strategies. The reported values are averaged over five repeated runs.}
    \label{fig:judge_strategy_ablation}
\end{figure}

The first conclusion is that a small amount of repeated voting is beneficial. Figure~\ref{fig:judge_strategy_ablation} shows that a single bidirectional judgment already correlates strongly with human reference scores, but moving to \textit{majority-3} further improves the overall correlation on the full validation set from 0.8543 to 0.8628. This gain suggests that limited repetition can suppress occasional judgment errors and yield more reliable semantic-equivalence decisions.

On the other hand, more voting is not necessarily better. Although \textit{majority-5} achieves the best result on GigaSpeech, its overall correlation on the full validation set is lower than that of \textit{majority-3}, and the same diminishing-return pattern appears for \textit{majority-7}. The extra rounds therefore add cost more reliably than they add quality. From a practical perspective, \textit{majority-3} offers the best trade-off between robustness and efficiency, making it a sensible default for the $S^2ER$ protocol.

\section{Conclusion}

In this work, we formulated \textbf{Interactive ASR} as a multi-turn semantic refinement problem. To address this setting, we proposed \textbf{Agentic ASR}, which combines single-pass ASR with LLM-based semantic correction, intent routing, and reasoning-based editing. We further introduced \textbf{$S^2ER$} to measure sentence-level semantic equivalence, and developed an \textbf{Interactive Simulation System} for scalable and reproducible multi-turn evaluation.

Experiments on multilingual, named-entity-intensive, and code-switching benchmarks show that iterative interaction consistently reduces semantic errors, with most of the benefit emerging in the first few rounds. Compared with WER, CER, NER, and MER, $S^2ER$ captures these gains more faithfully because the improvements are primarily semantic rather than merely lexical. Ablation studies further suggest that smaller LLMs remain usable in this framework, while stronger reasoners deliver more stable editing and better final performance.

Two directions appear especially promising for future work. First, richer interactive supervision, such as real user correction traces or automatically constructed interaction data, could be incorporated into training to improve robustness under realistic deployment conditions. Second, post-training a smaller task-specific refinement model is an attractive direction, since compact models already exhibit basic correction ability but still lag behind larger models in stability and precision.

\bibliographystyle{IEEEtran}
\bibliography{bibo}

\end{document}